\documentclass[conference,a4paper]{APSIPA2012}
\usepackage{multirow}
\usepackage{amsmath}
\usepackage[psamsfonts]{amssymb}
\usepackage{amsxtra}
\usepackage{threeparttable}
\usepackage{epsfig}
\usepackage{epstopdf,graphicx}
\usepackage{color}
\usepackage{epstopdf}
\usepackage{url}

\begin{document}

\title{Improved Deep Speaker Feature Learning for Text-Dependent Speaker Recognition}

\author{%
\authorblockN{%
Lantian Li, Yiye Lin, Zhiyong Zhang, Dong Wang$^*$
}
\authorblockA{%
Center for Speech and Language Technologies, Division of Technical Innovation and Development \\ Tsinghua National Laboratory for Information Science and Technology \\ Center for Speech and Language Technologies, Research Institute of Information Technology \\ Department of Computer Science and Technology, Tsinghua University, Beijing, China\\
\{lilt,lyy,zhangzy\}@cslt.riit.tsinghua.edu.cn; \\
wangdong99@mails.tsinghua.edu.cn
}
}

\maketitle
\thispagestyle{empty}

\begin{abstract}

A deep learning approach has been proposed recently to derive speaker identifies (d-vector) by
a deep neural network (DNN). This approach
has been applied to text-dependent speaker recognition tasks and shows reasonable performance gains when
combined with the conventional i-vector approach. Although promising, the existing d-vector
implementation still can not compete with the i-vector baseline.
This paper presents two improvements for the deep learning approach: a phone-dependent
DNN structure to normalize phone variation, and a new scoring approach based on
dynamic time warping (DTW).
Experiments on a text-dependent speaker recognition task demonstrated that the proposed methods can provide considerable performance improvement over the existing d-vector implementation.

\end{abstract}


\section{Introduction}

Most modern speaker recognition systems are based on human-crafted acoustic features,
for example the Mel frequency cepstral coefficients (MFCC).
A problem of the MFCC (and other primary features) is that it involves plethora information besides
the speaker identity, such as phone content, channels, noises, etc. These heterogeneous and noisy
information convolve together, making it difficult to be used for either speech
recognition or speaker recognition. All modern speaker recognition systems rely on a statical model
to `purify' the desired speaker information. For example, in the famous Gaussian mixture model-universal background model (GMM-UBM)
framework~\cite{ES2}, the acoustic space is divided into subspaces in the form of Gaussian components, and each
subspace roughly represents a phone. By formulating the speaker recognition task as
subtasks on the phone subspaces, the GMM-UBM model can largely eliminate the impact of phone content and
other acoustic factors. This idea is shared by many advanced techniques derived from GMM-UBM,
including the joint factor analysis (JFA)~\cite{ES3} and the i-vector model~\cite{ES4}.

In spite of the great success, the GMM-UBM approach and the related methods are still limited
 by the lack of discriminative capability of the acoustic features.
Some researchers proposed solutions based on discriminative models. For
example, the SVM approach for GMM-UBMs~\cite{ES5} and the PLDA approach for i-vectors~\cite{ES6}. All these
discriminative methods achieved remarkable success.
Another direction is to look for more task-oriented features, i.e., features that
are more discriminative for speaker recognition~\cite{ES7}. Although it seems to be
straightforward, this `feature engineering'
turns out to be highly difficult.
A major reason, in our mind, is that most of the proposed delicate features are human-crafted
and therefore tend to be fragile in practical usage.

Recent research on deep learning offers a new idea of `feature learning'. It has been
shown that with a deep neural network (DNN), task-oriented features can be learned
layer by layer from very raw input. For example in automatic speech recognition (ASR),
phone-discriminative features can be learned from spectra or filter bank energies (Fbanks).
This learned features are very powerful and have defeated the MFCC
that has dominated in ASR for several decades~\cite{ES8}.
This capability of DNNs in learning task-oriented features can be utilized to
learn speaker-discriminative features as well. A recent study shows that this is possible
at least on text-dependent tasks~\cite{ES1}. The authors reported
that reasonable performance can be achieved with the DNN-learned feature, and additional
performance gains can be obtained by combining the DNN-based approach and the
i-vector approach.

Although the DNN-based feature learning shows great potential, the existing implementation still
can not compete with the i-vector baseline. There are at least two drawbacks
with the current implementation: First, the DNN model does not use any information about the
phone content, which leads to difficulty when inferring speaker-discriminative features;
second, the evaluation (speaker scoring) is based on speaker vectors (so called `d-vectors'), which are derived by
averaging the frame-wise DNN features. This simple average ignores
the temporal constraint that is highly important for text-dependent tasks. Note that for
tasks with a fixed test phrase, the two drawbacks are closely linked to each other.

This paper follows the work
in~\cite{ES1} and provides two enhancements for the DNN-based feature learning:
First, phone posteriors are involved in the DNN input so that
speaker-discriminative features can be learned easier by alleviating the
impact of phone variation; second, two scoring methods that consider the temporal
constraint are proposed: segmentation pooling and dynamic time warping (DTW)~\cite{ES11}.

The rest of the paper is organized as follows. Section~\ref{sec:rel} describes some
related work, and Section~\ref{sec:theory} presents the DNN-based feature learning. The
new methods are proposed in Section~\ref{sec:improve} and the experiments are presented in
Section~\ref{sec:exp}. Finally, Section~\ref{sec:conl}
concludes this paper and discusses some future work.

\section{Related work}
\label{sec:rel}

This paper follows the work in~\cite{ES1} and provides several extensions. Particularly,
the speaker identity in~\cite{ES1} is represented by a d-vector derived by average pooling,
which is quite neat and efficient, but loses much information of the test signal,
such as the distributional property and the temporal constraint. One of the main contribution
of this paper is to investigate how to utilize the temporal constraint in the DNN-based approach.

The DNN model has been studied in speaker recognition in several ways. For example,
in~\cite{ES9}, DNNs trained for ASR were used to replace the UBM model to derive
the acoustic statistics for i-vector models. In~\cite{ES10}, a DNN was used to
replace PLDA to improve discriminative capability of i-vectors. All these methods
rely on the generative framework, i.e., the i-vector model. The DNN-based
feature learning presented in this paper is purely discriminative, without
any generative model involved.

\section{DNN-based feature learning}
\label{sec:theory}

It is well-known that DNNs can learn task-oriented features from raw input
layer by layer. This property has been employed in ASR where
phone-discriminative features are learned from very low-level
features such as Fbanks or even spectra~\cite{ES8}. It has been shown
that with a well-trained DNN, variations irrelevant to the learning
task can be gradually eliminated when the feature
propagates through the DNN structure layer by layer.
This feature learning is so powerful that in ASR, the primary Fbank feature has
defeated the MFCC feature that was carefully designed by people and
dominated in ASR for several decades.

This property can be also employed to learn speaker-discriminative features. Actually
researchers have put much effort in searching for features that are more discriminative
for speakers~\cite{ES7}, but the effort is mostly vain and the MFCC is still
the most popular choice. The success
of DNNs in ASR suggests a new direction, that speaker-discriminative
features can be learned from data instead of being crafted by hand. The learning can be
easily done and the process is rather similar as in ASR, with the only difference
that in speaker recognition, the learning goal
is to discriminate different speakers.

\begin{figure}[htb]
   \centering
   \includegraphics[width=\linewidth]{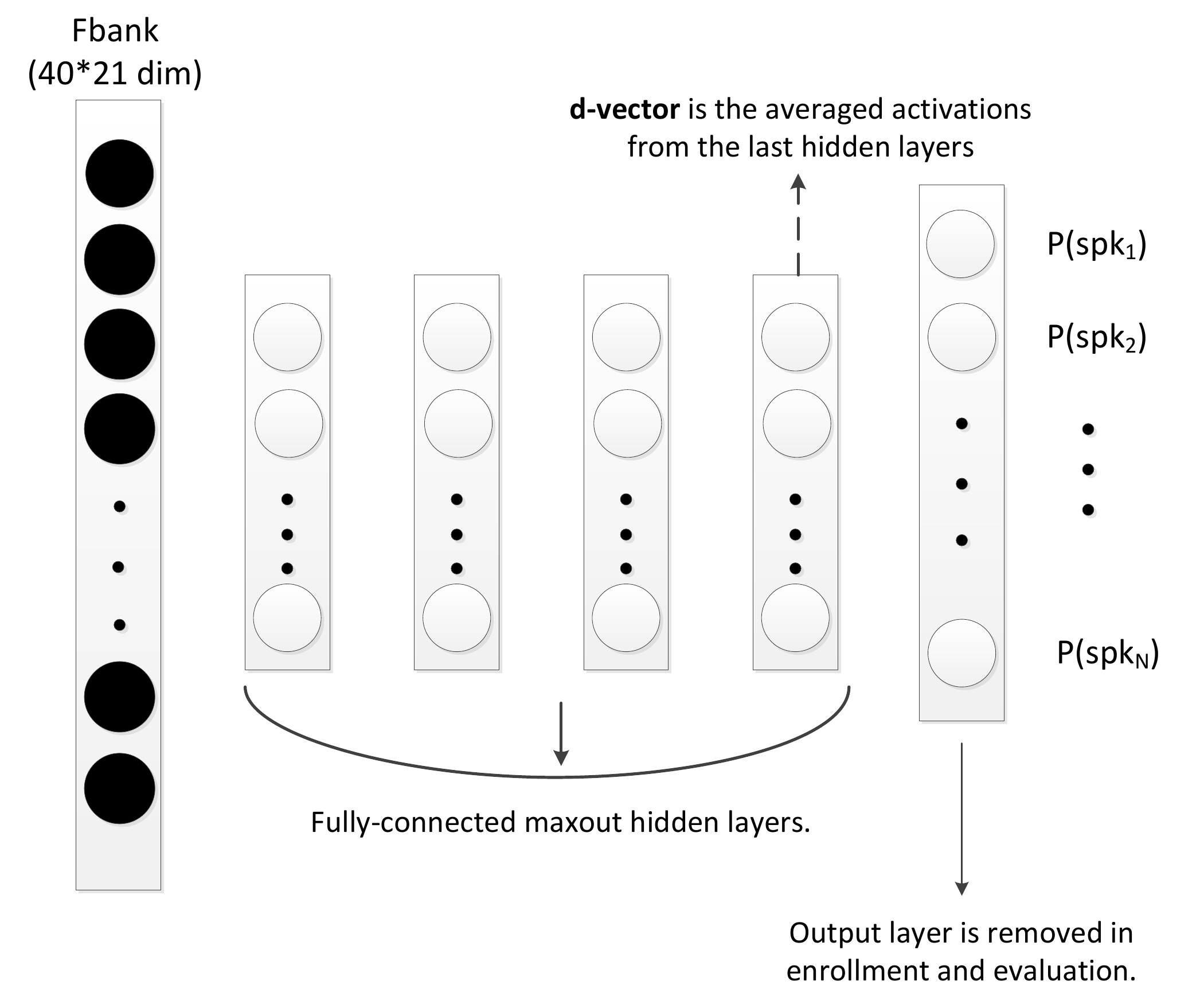}
   \caption{The DNN structure used for learning speaker-discriminative features.}
   \label{fig:dnn}
\end{figure}

Fig.~\ref{fig:dnn} presents the DNN structure used in this work for speaker-discriminative feature learning.
Following the convention of ASR, the input layer involves a window of 20-dimensional Fbanks.
The window size is set to $21$, which was found to be optimal in our work.
The DNN structure involves $4$ hidden layers, and each consists of $200$
units. The units of the output layer correspond to the speakers in the training data, and the number is $80$ in our experiment.
The 1-hot encoding scheme is used to label the target, and the training
criterion is set to cross entropy. The learning rate is set to $0.008$ at the beginning, and is
halved whenever no improvement on a cross-validation (CV) set is found. The training process stops when the learning
rate is too small and the improvement on the CV set is too marginal. Once the DNN has been trained
successfully, the speaker-discriminative features can be read from
the last hidden layer.

In the test phase, the features are extracted for all frames of the given utterance. To derive utterance-based representations,
an average pooling approach was used in~\cite{ES1}, where the frame-level features are
averaged and the resultant vector is used to represent the speaker. This vector is called `d-vector' in~\cite{ES1},
and we adopt this name in this work. The same methods used for i-vectors can be used for d-vectors to conduct
the test, for example by computing the cosine distance or the PLDA score.

\subsection{Comparison between i-vectors and d-vectors}

The two kinds of speaker vectors, the d-vector and the i-vector, are fundamentally different.
I-vectors are based on a linear Gaussian model, for which the learning is unsupervised
and the learning criterion is maximum likelihood on acoustic features; in contrast, d-vectors are
based on neural networks, for which the learning is supervised, and the learning criterion is maximum
discrimination for speakers. This difference leads to several advantages with d-vectors: First,
it is a `discriminative' vector, which represents speakers by removing speaker-irrelevant variance,
and so sensitive to speakers and invariant to other disturbance;
second, it is a `local' speaker description that uses only local context, so can be inferred from very short
utterances; third, it relies on `universal' data to learn the DNN model, which makes it
possible to learn from large amounts of data that are task-independent.

\section{Improved deep feature learning}
\label{sec:improve}

There are several limitations in the implementation of the feature learning paradigm presented
in the previous section. First, it does not involve any prior knowledge
in model training, for example phone identities. Second, the simple average pooling does not consider
the temporal information which is particularly important for text-dependent recognition tasks.
Several approaches are proposed in this section to address these problems.

\begin{figure}[htb]
   \centering
   \includegraphics[width=\linewidth]{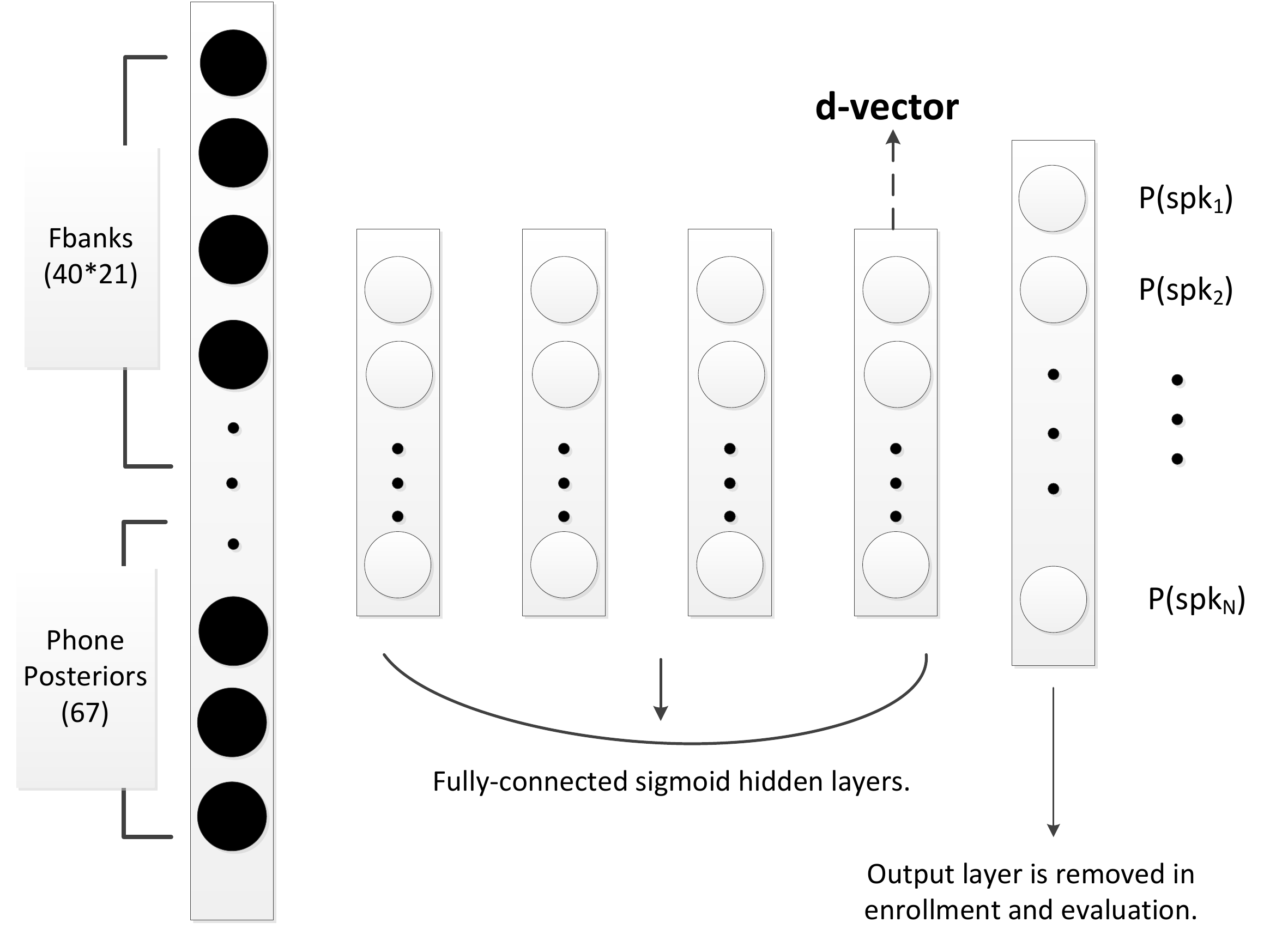}
   \caption{Phone-dependent DNN structure used for learning speaker-discriminative features.}
   \label{fig:pt-dnn}
\end{figure}

\subsection{Phone-dependent training}

A potential problem of the DNN-based feature learning described in the previous section
is that it is a `blind learning', i.e., the features are learned from raw data
without any prior knowledge. This means that the learning purely relies on the complex deep structure
of the DNN model and a large amount of data to discover speaker-discriminative patterns. If the training data is
abundant, this is often not a problem; however in tasks with a limited amount of data,
for instance the text-dependent task in our hand, this blind learning tends to be
difficult because there are too many speaker-irrelevant variations involved in the raw data, particularly phone contents.

A possible solution is to supply the DNN model extra information about which phone is spoken at each frame.
This can be simply achieved
by adding a phone indicator in the DNN input. However, it is often not easy to get the phone alignment
in practice. An alternative way is to supply a vector of phone posterior probabilities for each frame, which is
a `soft' phone alignment and can be easily obtained from a phone-discriminative model. In this work, we choose
to use a DNN model that was trained for ASR to produce the phone posteriors. Fig.~\ref{fig:dnn} illustrates
how the phone posteriors are involved in the DNN structure. The training process
does not change for the new structure.

\subsection{Segment pooling and dynamic time warping}

Text-dependent speaker recognition is essentially a sequential pattern matching problem,
but the current d-vector approach derives speaker identities
as single vectors by average pooling, and then formulates speaker recognition as vector matching.
This is certainly not ideal as the temporal constraint is totally ignored when
deriving the speaker vector. A possible solution is to segment an enrollment/test utterance
into several pieces, and derive the speaker vector for each piece. The speaker identity of the utterance
is then represented by the sequence of the piece-wise speaker vectors, and speaker matching is
conducted by matching the corresponding vector sequences. This paper adopts a simple
sequence matching approach: the two sequences are assumed to be identical in length, and
the matching is conducted piece by piece independently. Finally the matching score takes the average
of scores on all the pieces.

Note that for the i-vector approach, this segmentation
method is not feasible, since i-vectors are inferred from feature distributions and so the
piece-wised solution simply degrades the quality of the i-vector of each piece. For d-vectors,
this approach is totally fine as they are inferred from local context and the segmentation
does not impact quality of the piece-wise d-vectors very much.

A more theoretical treatment is based on dynamic time warping (DTW)~\cite{ES2}.
The DTW algorithm is a principle way to measure similarities between two variable-length temporal sequences.
In the most simple sense, DTW searches for an optimal path that matches two sequences with the lowest cost,
by employing the dynamic programming (DP) method to reduce the search complexity. In our task,
the DNN-extracted features of an utterance are treated as a temporal sequence. In test,
the sequence derived from the enrollment utterance and the sequence derived from the test utterance are
matched by DTW, where the cosine distance is used to measure the similarity between two frame-level
DNN features. Principally, segment pooling can be regarded as a special case of DTW, where
the two sequences are in the same length, and the matching between the two sequences is piece-wise.

\section{Experiments}
\label{sec:exp}

\subsection{Database}

The experiments are performed on a database that involves a limited set of short phrases. The entire
database contains recordings of $10$ short phrases from $100$ speakers (gender balanced), and each phrase
contains $2\sim5$ Chinese characters. For each speaker, every phrase is recorded $15$ times,
amounting to $150$ utterances per speaker.

The training set involves $80$ randomly selected speakers, which results in $12000$ utterances in total.
To prevent over-fitting, a cross-validation (CV) set containing $1000$ utterances is selected
from the training data, and the remaining $11000$ utterances are used for model training, including the
DNN model in the d-vector approach, and the UBM, the T matrix, the LDA and PLDA model in the i-vector approach.

The evaluation set consists of the remaining $20$ speakers. The evaluation is
performed for each particular phrase. For each phrase, there are $44850$ trails, including
$2100$ target trails and $42750$ non-target trials.
For the sake of neat presentation, we report the results with $5$ short phrases and
use `Pn' to denote the n-th phrase. The conclusions obtained here
generalize well to other phrases.

\subsection{Baseline}

Two baseline systems are built, one is based on i-vectors and the other is based on d-vectors. The
acoustic features of the i-vector system are $39$-dimensional MFCCs, which consist of
$13$ static components (including C0) and the first- and second-order derivatives. The number of
Gaussian components of the UBM is $128$, and the dimension of the i-vector is $200$.
The d-vector baseline uses the DNN structure shown in Fig.~\ref{fig:dnn}. The average pooling
 is used to derive d-vectors.
The acoustic features are $40$-dimensional Fbanks, with the
left and right $10$ frames concatenated together.
The frame-level features are extracted from the last hidden layer, and the dimension is $200$.

Table~\ref{tab:baseline} presents the results in terms of equal error rate (EER). It can be seen
that the i-vector system generally outperforms the d-vector system in a significant way. Particularly,
the discriminative methods (LDA and PLDA) clearly improves the i-vector system, however for
the d-vector system, no improvement was found by these methods. This is not surprising, since the
d-vectors have been discriminative by themselves. For this reason, LDA and PLDA are not considered
any more for d-vectors in the following experiments.


\begin{table}[th]
        \centering
          \caption{Performance of baseline systems}
          \label{tab:baseline}
          \begin{tabular}{l|c|c|c|c}
            \hline
                &       &\multicolumn{3}{|c}{EER\%} \\
           \hline
                &Phrase & cosine & LDA & PLDA \\
            \hline
          i-vector  &P1 & 2.86  & 1.81 & 1.71\\
                    &P2 & 1.52  & 2.29 & 1.57 \\
                    &P3 & 3.43  & 3.05 & 3.05  \\
                    &P4 & 3.19  & 2.86 & 2.71 \\
                    &P5 & 3.57  & 3.00 & 2.67 \\
            \hline

          d-vector  &P1 & 10.29  & 9.81 & 12.67 \\
                    &P2 & 10.52  & 10.57 & 12.29 \\
                    &P3 & 10.10  & 9.33 & 10.48 \\
                    &P4 & 10.38  & 9.95 & 11.10 \\
                    &P5 & 9.14  & 9.29 & 11.10 \\
          \hline
          \end{tabular}
\end{table}

\subsection{phone-dependent learning}

In this experiment, the phone posteriors are included in the DNN input, as shown in Fig.~\ref{fig:pt-dnn}.
The phone posteriors are produced by a DNN model that was trained for ASR with a Chinese database consisting of $6000$
hours of speech data. The phone set consists of $66$ toneless initial and finals in Chinese, plus the silence phone.
The results are shown in the second row of Table~\ref{tab:seg}, denoted by `DNN+PT'.
It can be seen that the phone-dependent training leads to
marginal but consistent performance improvement for the d-vector system.

\begin{table}[htb]
        \centering
          \caption{Performance with improved DNN feature learning}
          \label{tab:seg}
          \begin{tabular}{l|c|c|c|c|c}
            \hline
                         &\multicolumn{5}{|c}{EER\%}\\
            \hline
                                     &  P1 &  P2 & P3 & P4 & P5 \\
            \hline
                DNN baseline         & 10.29 & 10.52 & 10.10 & 10.38 & 9.14 \\
                DNN+PT               & 10.24 & 10.05 & 9.81 & 9.48 & 8.71 \\
            \hline
                DNN+PT+SEG          & 9.95  & 8.90 & 8.95 & 9.76 & 8.67 \\
                DNN+PT+DTW           & 9.14  & 8.38 & 8.52 & 8.86 & 8.14 \\
            \hline
          \end{tabular}
\end{table}

\subsection{Segment pooling and DTW}

\begin{figure}[htb]
   \centering
   \includegraphics[width=\linewidth]{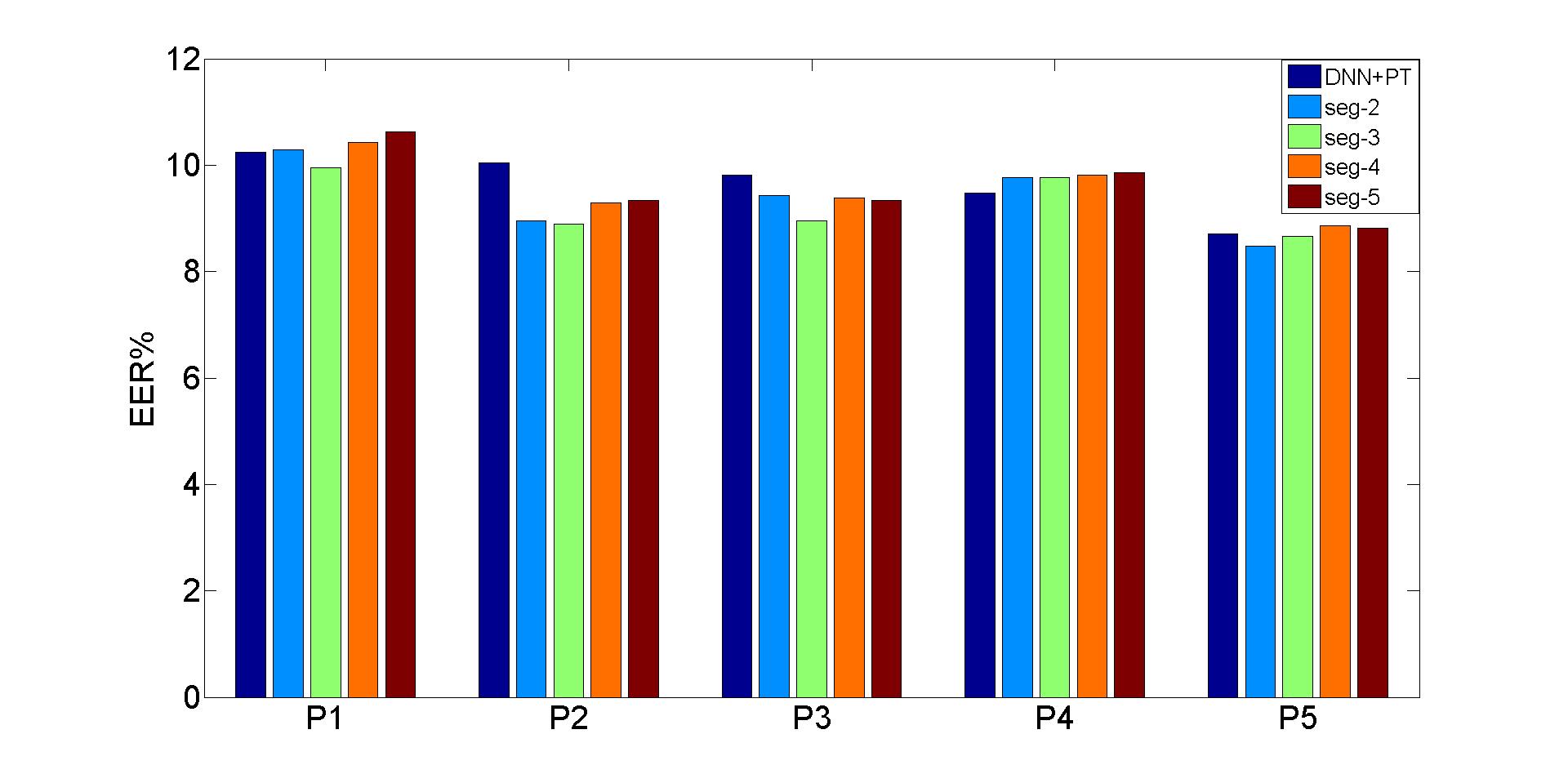}
   \caption{{\it EER results with segment pooling.}}
   \label{fig:seg}
\end{figure}

As presented in Section~\ref{sec:improve}, the segment pooling approach segments an input
utterance into $n$ pieces,
and derives a d-vector for each piece. The scoring is conducted on the piece-wise d-vectors
independently, and the average of scores on these pieces is taken as the utterance-level score.
The results are shown in Fig.~\ref{fig:seg}, where seg-$n$ means that each utterance is
segmented into $n$ pieces. It can be seen that segment pooling offers clear performance
improvement.

For a more clear comparison, the EER results with $3$ segments are shown in
Table~\ref{tab:seg}, denoted by `DNN+PT+SEG'.
It is clear to see that by the segmentation, significant performance is obtained.

The DTW results are shown in the fourth row of Table~\ref{tab:seg}, denoted by `DNN+PT+DTW'. It can be
observed that DTW generally outperforms segment pooling.

\begin{table}[!htbp]
        \centering
          \caption{Performance of system combination}
          \label{tab:fusion}
          \begin{tabular}{l|c|c|c|c|c}
            \hline
                         &\multicolumn{5}{|c}{EER\%}\\
            \hline
                                     &  P1 &  P2 & P3 & P4 & P5 \\
            \hline
                PLDA             & 1.71 & 1.57 & 3.05 & 2.71 & 2.67 \\
                DNN+PT+DTW       & 9.14 & 8.38 & 8.52 & 8.86 & 8.14 \\
            \hline
                Combination           & 1.52 & 1.38 & 2.33 & 2.33 & 2.38 \\
            \hline
          \end{tabular}
\end{table}

\subsection{System combination}

Following~\cite{ES1}, we combine the best i-vector system (PLDA) and the best d-vector system (DNN+PT+DTW).
The combination is simply done by interpolating the scores obtained from the two systems: $\alpha s_{iv} + (1-\alpha) s_{dv}$, where
$s_{iv}$ and $s_{dv}$ are scores from the i-vector and d-vector systems respectively, and $\alpha$ is the interpolation factor. The EER results with the
optimal $\alpha$ are shown in Table~\ref{tab:fusion}.
It can be seen that the combination leads to the best performance we can obtain so far.

\section{Conclusions}
\label{sec:conl}
This paper presented several enhancements for the DNN-based feature learning approach in speaker recognition. We presented a phone-dependent DNN model to supply phonetic information when learning speaker features, and proposed two scoring methods based on a segment pooling and DTW respectively
to leverage temporal constraints. These extensions significantly improved
performance of the d-vector system. Future work involves investigating more
complicated statistical models for d-vectors,
and use large amounts of data to learn more powerful speaker-discriminative features.


\bibliographystyle{IEEEtran}
\small{
\bibliography{segvector}
}

\end{document}